\documentclass[10pt,twocolumn,letterpaper]{article}

\usepackage{wacv}              

\usepackage{graphicx}
\usepackage{amsmath}
\usepackage{amssymb}
\usepackage{booktabs}
\usepackage[graphicx]{realboxes}

\usepackage[pagebackref,breaklinks,colorlinks]{hyperref}

\usepackage[capitalize]{cleveref}
\crefname{section}{Sec.}{Secs.}
\Crefname{section}{Section}{Sections}
\Crefname{table}{Table}{Tables}
\crefname{table}{Tab.}{Tabs.}

\def\wacvPaperID{802} 
\def\confName{WACV}
\def\confYear{2024}

\begin{document}

\title{Reverse Knowledge Distillation: Training a Large Model using a Small One for Retinal Image Matching on Limited Data}
\author{Sahar Almahfouz Nasser*\\
Indian Institute of Technology Bombay\\
Mumbai, Maharashtra, India\\
{\tt\small 194072001@iitb.ac.in}
\and
Nihar Gupte*\\
Indian Institute of Technology Bombay\\
Mumbai, Maharashtra, India\\
{\tt\small 213070002@iitb.ac.in}
\and
Amit Sethi\\
Indian Institute of Technology Bombay\\
Mumbai, Maharashtra, India\\
{\tt\small asethi@iitb.ac.in}
}

\maketitle
\footnote{* Indicates equal contribution}
\begin{abstract}
Retinal image matching plays a crucial role in monitoring disease progression and treatment response. However, datasets with matched keypoints between temporally separated pairs of images are not available in abundance to train transformer-based model. We propose a novel approach based on reverse knowledge distillation to train large models with limited data while preventing overfitting. Firstly, we propose architectural modifications to a CNN-based semi-supervised method called SuperRetina~\cite{liu2022semi} that help us improve its results on a publicly available dataset. Then, we train a computationally heavier model based on a vision transformer encoder using the lighter CNN-based model, which is counter-intuitive in the field knowledge-distillation research where training lighter models based on heavier ones is the norm. Surprisingly, such reverse knowledge distillation improves generalization even further. Our experiments suggest that high-dimensional fitting in representation space may prevent overfitting unlike training directly to match the final output. We also provide a public dataset with annotations for retinal image keypoint detection and matching to help the research community develop algorithms for retinal image applications. 

\end{abstract}

\section{Introduction}
\label{sec:intro}

Keypoint detection and matching, also referred to as feature point extraction or feature detection, is a fundamental task in the field of computer vision. Its primary objective is to identify and locate prominent points or landmarks within images. These keypoints possess distinct and robust characteristics, making them valuable for various applications including object recognition, image registration, image stitching, pose estimation, facial recognition, and augmented reality. Typically, keypoint detection algorithms identify points in an image that are locally unique in intensity, color, or texture. These points are also expected to be invariant to changes in scale, rotation, and illumination. Keypoint locations are typically represented as 2D or 3D coordinates. Along with location, a feature descriptor is extracted for each point, which helps in its recognition or matching with corresponding points in another image. 

Throughout the years, numerous methods have been developed for keypoint detection. These methods span from classical techniques like the Harris corner detector~\cite{ryu2011formula}, scale-invariant feature transform (SIFT)~\cite{lowe2004distinctive}, and speeded-up robust features (SURF)~\cite{bay2006surf}, to more recent approaches based on deep learning, such as oriented fast and rotated BRIEF (ORB)~\cite{rublee2011orb} and SuperPoint~\cite{detone2018superpoint}.

Several methods have been proposed for keypoints matching in retinal images. Addison et al. \cite{addison2015low} introduced a technique called low dimensional step pattern analysis (LoSAP) for image registration. LoSAP is capable of handling intensity changes and is invariant to rotation. However, the SPA descriptor used in LoSAP lacks discriminative power in identifying specific eye identities. Truong et al. \cite{truong2019glampoints} presented a semi-supervised CNN-based feature point detector known as Greedily Learned Accurate Match Points (GLAMpoints) specifically designed for matching and registering retinal images. GLAMpoints utilizes deep learning techniques to improve the accuracy and precision of keypoint matching. Another approach, proposed by Hernandez et al. \cite{hernandez2020rempe}, involves a registration framework based on eye modeling. This framework simultaneously estimates eye pose and shape and solves the registration problem as a 3D pose estimation task, utilizing corresponding points in the retinal images.

Our work makes the following contributions. Firstly, we are releasing annotations on a curated dataset specifically for detecting keypoints in retinal images. Additionally, we introduced a more powerful architecture that surpasses the state-of-the-art SuperRetina model for detecting keypoints in retinal images~\cite{liu2022semi}. Furthermore, we investigated the significance of using a CNN versus incorporating a transformer model. Finally, we propose and explore a reverse knowledge distillation as a training methodology for large models, especially when limited training data is available. In this approach, the large model (student) learns from the small model (teacher) to enhance its performance and surpass the performance of its teacher as we will see later in section~\ref{sec:experiments}.

\section{Related work}
\label{sec:related}

In this section, we will review important methods for keypoint detection. We will also provide an overview of vision transformers, emphasizing their significance in computer vision. Additionally, we will delve into various techniques employed for training vision transformers when data is limited.

\subsection{Keypoint detection}

Traditional keypoint detection algorithms, such as Harris corner detector \cite{ryu2011formula}, SIFT \cite{lowe2004distinctive} and SURF \cite{bay2006surf}, have been widely used in computer vision applications for decades. These algorithms detect keypoints in images, which are invariant to scaling, rotation, and lighting changes. They then describe the local image patch around the keypoints using a set of features, which can be used to match keypoints between different images or recognize objects. However, these techniques have some drawbacks, including high computational cost, limited accuracy under extreme lighting and viewpoint changes, and difficulty in handling occlusions and cluttered backgrounds. 

Deep learning-based keypoint detection algorithms, which can automatically learn robust and discriminative features directly from data, and can handle complex and diverse image variations more effectively, have been proposed in recent years. These methods have shown promising results in various applications, such as object detection, semantic segmentation, and image retrieval.

In deep learning, there are different types of keypoint detection algorithms: supervised, semi-supervised, self-supervised and unsupervised techniques. Supervised techniques require annotated data, where the keypoints are manually labeled in the training images. These algorithms are useful in applications where there is a large amount of labeled data available, such as facial recognition or object detection. On the other hand, unsupervised techniques do not require labeled data, and instead, the network learns to detect keypoints by maximizing certain objectives, such as the amount of information preserved during feature extraction. These methods are useful in applications where labeled data is scarce or expensive to obtain, such as medical imaging or remote sensing.

Prominent deep-learning based keypoint detection methods include UnsuperPoint, SuperPoint, GLAMpoints, and SuperRetina. UnsuperPoint \cite{christiansen2019unsuperpoint} uses a new unsupervised training approach based on a combination of a differentiable soft nearest neighbor loss and an unsupervised clustering loss. SuperPoint \cite{detone2018superpoint} is a self-supervised deep learning-based algorithm for keypoint detection and description. It uses a novel loss function to train on unannotated images, making it more scalable and adaptable to various applications. The loss functions (geometric consistency loss, and descriptor matching loss) encourage the network to learn to predict the spatial location of the keypoints and their descriptors without supervision. It uses a convolutional neural network (CNN) to extract keypoint locations and descriptors from images.
The main difference between UnsuperPoint and SuperPoint is the training process. SuperPoint is trained in a self-supervised manner, whereas UnsuperPoint is trained in an unsupervised manner. Additionally, UnsuperPoint achieves state-of-the-art performance on various benchmarks and outperforms SuperPoint in some challenging scenarios, such as large viewpoint changes and illumination variations. GLAMpoints \cite{truong2019glampoints} is a semi-supervised deep learning-based algorithm for interest point detection and description. It uses a novel greedy training strategy to learn keypoint detection and description in an end-to-end manner and learns to select the most accurate keypoints and their descriptors resulting in high accuracy and efficiency. GLAMpoints outperforms SuperPoint and UnsuperPoint in terms of accuracy and efficiency, especially under challenging scenarios such as large viewpoint changes, scaling, and rotation, and on various benchmarks such as HPatches~\cite{balntas2017hpatches}. Moreover, GLAMpoints is designed to handle multiple object instances in the same image, making it suitable for multi-object tracking and matching. SuperRetina \cite{liu2022semi} is a semi-supervised approach for keypoint detection and description in retinal images. The method uses a combination of labeled and unlabeled data to improve the performance of the keypoint detector and descriptor. The approach consists of three main components: a supervised keypoint detector, an unsupervised keypoint descriptor, and a semi-supervised loss function that combines both labeled and unlabeled data. Their proposed method uses an iterative refinement process to improve the accuracy and robustness of the keypoint matches. The refinement process involves removing outlier matches and adding new matches based on geometric constraints.

\subsection{Vision transformers}

Taking inspiration from the success of transformer models in natural language processing, vision transformers adopt the same self-attention mechanism to process visual data~\cite{liu2021swin,yu2022metaformer}. By treating the entire image as a sequence of tokens or patches, vision transformers excel at capturing global dependencies and long-range interactions within the image. They have demonstrated promising performance across various tasks, including image classification, object detection, and segmentation.  Vision transformers typically require a substantial amount of labeled training data to acquire meaningful representations and achieve good generalization on new examples. Consequently, training vision transformers with limited data can be challenging due to their extensive parameter count and the risk of overfitting. 

Several strategies can be employed to mitigate the challenge of training in smaller datasets~\cite{steiner2021train}. One effective approach is data augmentation, which involves generating synthetic training examples by applying transformations like rotations, translations, flips, and color variations to the available data~\cite{steiner2021train}. This augmentation technique expands the effective size of the training set and aids the model in learning more robust and invariant features. Another strategy is transfer learning, whereby a pre-trained vision transformer model on a large-scale dataset (such as ImageNet) serves as a starting point~\cite{weiss2016survey}. The model can then be fine-tuned on the limited dataset of interest. Transfer learning allows the model to leverage the knowledge acquired from the larger dataset, thereby enhancing its generalization even with limited data. Furthermore, regularization techniques like dropout and weight decay can be employed to prevent overfitting and improve generalization~\cite{steiner2021train}. These regularization techniques encourage the model to avoid excessive reliance on specific training examples, enabling it to learn more general patterns and representations.

\subsection{Knowledge distillation}

Knowledge distillation~\cite{habib2023knowledge} is a technique whereby a pre-trained model (known as the teacher model) is utilized to guide the training of another model (the student model). The student model learns to mimic the predictions or internal data representations (features) of the teacher model. This is typically done to leverage the knowledge and generalization capabilities of the larger teacher model into a smaller student model. 

In our work, we adopt distillation in the reverse direction where a small CNN-based model serves as the teacher model, while a large transformer-based model acts as the student model. We hypothesize that larger models can overfit when they fit a smaller-dimensional output, but this curse can be broken if they are trained to fit a larger dimensional representation (feature vector).

\section{Datasets}
\label{sec:dataset}

To train a network that can detect keypoints on retinal images, it is essential to have images with accurately labeled keypoints. Previous methods used the FIRE dataset for testing the detection of keypoints on retinal images, and to ensure a fair comparison with those methods, we adopted the same approach and exclusively used the FIRE dataset for testing in our study. Unfortunately, despite reaching out to others who had utilized private datasets for training their networks, we were unable to obtain access to those datasets.

Consequently, we decided to create our own training annotation set using publicly available datasets originally intended for other tasks, such as retinal disease classification. As we lack expertise in the domain, we restricted our dataset exclusively to normal images, with the hope that it would still help develop the capability to detect keypoints in abnormal images during the testing phase, including specific images present in the FIRE dataset. In this section we will provide detailed insights into our dataset and its annotations, and the FIRE dataset.

\subsection{Our Dataset}

\begin{figure}
  \centering
  \includegraphics[width=1.0\columnwidth]{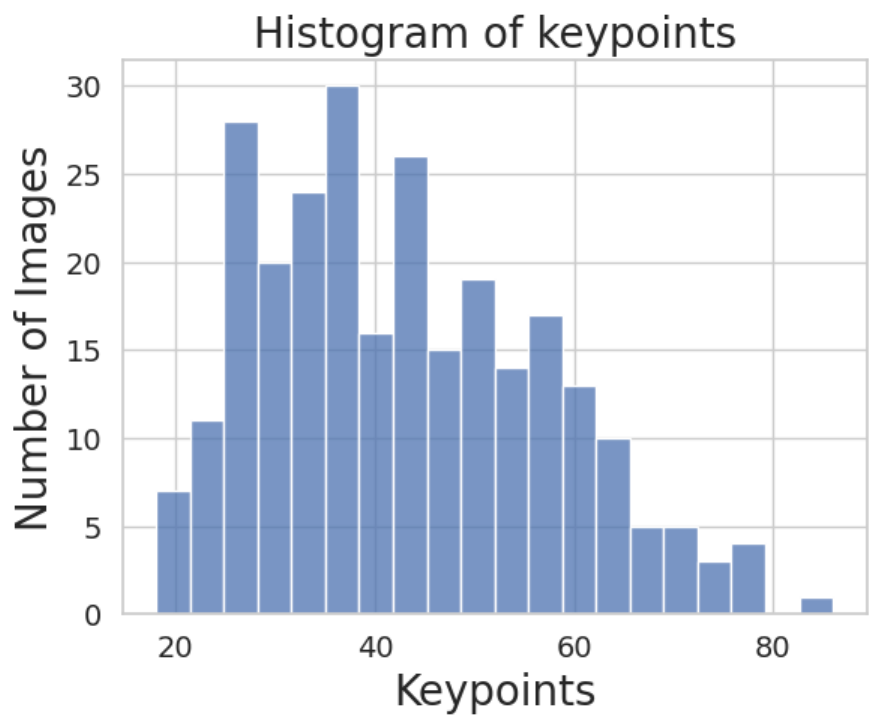}
  \caption{Distribution of keypoints per image in MeDAL dataset.}
  \label{dist}
\end{figure}

Our dataset comprises a total of 261 retinal images, which were divided into 208 images for training purposes and 61 images for validation~\cite{medal2023}. Each image underwent meticulous annotation to identify keypoints at intersections, crossovers, and bifurcations. The number of keypoints detected in each image varied from 18 to 86, with an average of 42.96 ± 14.03 keypoints. The distribution of keypoints across the images is visually represented in Figure \ref{dist}.

To assemble our dataset, we gathered images from two distinct sources. Firstly, we included 201 normal images obtained from the e-ophtha dataset \cite{e-ophtha}. Additionally, we incorporated 60 images from the retinal disease classification dataset \cite{RetinalDisease}, see the figure on the dataset provided in the supplemental material~\cite{Authors14}.
The annotation process was executed by a team of five engineering students, with each student assigned a specific set of images. On average, annotating a single image required approximately five minutes, while the annotation of a pair of images took approximately 8 minutes. For annotating the images, we developed a Python script to aid the process.

In Section \ref{sec:proposed}, we will observe that in order to utilize Swin UNETR~\cite{hatamizadeh2021swin} as the backbone of our network, we needed training a Swin UNETR in a self-supervised manner before using it and for that a substantial dataset was required. Since the distribution of retinal images significantly differs from that of the ImageNet data~\cite{deng2009imagenet}, we made the decision to gather a substantial retinal dataset specifically for training Swin UNETR. The dataset, comprising $\sim 1.9 K$ images, was collected from various online resources~\cite{KaggleDiabeticRetinopathy,KaggleGlaucomaDatasets, sarhan2021transfer,hernandez2017fire,RetinalDiseaseClassification} by our team. Additionally, this dataset was utilized to train the descriptor decoders as well.

For preprocessing, our initial step involves normalizing the images using z-score normalization. Subsequently, we applied contrast limited adaptive histogram equalization (CLAHE) and gamma correction. Finally, we divided the preprocessed images by 255. Throughout all our experiments, we consistently utilized the green channel, as it consistently demonstrated the most optimal performance.


\subsection{FIRE Dataset}

The FIRE dataset, which focuses on fundus image registration, consists of a total of 129 retinal images~\cite{hernandez2017fire}. These images were organized into 134 pairs based on the degree of overlap and deformation between them, with each pair belonging to a specific category. The dataset is divided into three categories: S, P, and A. The S category comprises 71 pairs of images with a large overlap ($>75\%$) and minimal visual anatomical differences. These pairs exhibit changes in brightness, as well as slight shifts and/or rotations. The P category contains 49 pairs of images with smaller overlaps compared to the S category. These pairs exhibit significant differences, including large shifts and rotations, between the images. The A category consists of 14 pairs of images with a large overlap. However, the images in each pair were acquired at different examinations, resulting in substantial anatomical changes such as spots, cotton-wool patches, and increased vessel tortuosity.

The retinal images were captured using a Nidek AFC-210 fundus camera, providing images with a resolution of $2912 \times 2912$ pixels and a field of view (FOV) of 45° in both the x and y dimensions. The images were obtained from 39 patients at the Papageorgiou Hospital, Aristotle University of Thessaloniki, Thessaloniki.

The figure on the dataset provided in the supplemental material~\cite{Authors14} illustrates examples from both our dataset~\cite{medal2023} and the FIRE dataset. The first and second rows display images from the e-ophtha dataset and the retinal disease classification dataset, respectively, along with our annotations presented as keypoints overlaid on these images. The third row showcases images from the FIRE dataset, accompanied by the corresponding annotations for reference.

\section{Proposed Method}
\label{sec:proposed}

SuperRetina \cite{liu2022semi} is a state-of-the-art (SOTA) technique for identifying informative keypoints in retinal images. It is an adapted version of the SuperPoint model~\cite{detone2018superpoint}, specifically designed to excel in retinal image analysis. This innovative approach leverages a semi-supervised learning framework, effectively combining supervised and unsupervised techniques to maximize the utilization of the limited amount of labeled data found in retinal image datasets. The network architecture comprises an encoder responsible for extracting downsampled feature maps from the input image, along with two decoders: one for detecting keypoints and another for generating descriptors associated with these keypoints. The keypoint detector is trained using a combination of labeled and unlabeled data, while the descriptor is trained using self-supervised learning methods. Rigorous experimentation conducted on established benchmark retinal image datasets showcases the proposed approach's exceptional performance in terms of keypoint detection and matching accuracy, outperforming existing methods~\cite{fire_dataset}. By employing a semi-supervised framework, this method effectively capitalizes on the available data, which is particularly advantageous in the domain of retinal image analysis where annotated data is often scarce. Consequently, the suggested method exhibits significant potential for enhancing various applications in retinal image analysis, including image registration, image alignment, and image-based disease diagnosis.

\subsection{UNet-empowered SuperRetina}
The initial encoder-decoder architecture of SuperRetina drew inspiration from the U-Net architecture \cite{ronneberger2015u}. In SuperRetina, the encoder is shallow, comprising a single convolutional layer followed by three convolutional blocks. Each block consists of two convolutional layers, a $2 \times 2$ max pooling layer, and a ReLU activation function. The decoder for keypoint detection includes three convolutional blocks with two convolutional layers in each block. It incorporates bilinear upsampling, ReLU activation, and concatenation, benefiting from skip connections originating from the encoder. Subsequently, feature maps of the same size as the input image are obtained, and a convolutional block with three convolutional layers and a sigmoid activation is employed to generate the detection map ($P$). 
For the descriptor decoder, the feature maps from the encoder are downsampled into more compact $\frac{w}{16} \times \frac{h}{16} \times d$ feature maps. Then, a transposed convolutional block is utilized to upsample the feature maps to match the size of the input image, ultimately producing a full-sized descriptor tensor ($D$) of dimensions $h \times w \times d$. All descriptors are l2-normalized.

Our focus lies in enhancing the performance of SuperRetina through improvements in the encoder design. To achieve this, we introduced several architectural modifications to SuperRetina. These modifications use two methods: a CNN-based approach and a transformer-based approach, aiming to improve the overall results.

\subsection{Large kernel-empowered SuperRetina}
Taking inspiration from the work of Jia et al. \cite{jia2022u}, which successfully enhances the performance of a basic U-Net architecture to rival that of the powerful transformer architecture, we propose a straightforward technique. This technique involves introducing kernels of varying sizes in each layer of the encoder of SuperRetina, aiming to effectively capture long-range dependencies in retinal image matching tasks.

In our approach, we made specific modifications to the architecture of SuperRetina's encoder. Instead of using a $3 \times 3$ kernel in each layer, we replaced it with three kernels of different sizes: $1 \times 1$, $3 \times 3$, and $5 \times 5$. By incorporating these modifications, the modified SuperRetina surpasses the state-of-the-art method for retinal image matching. It surpasses all previous methods evaluated on the FIRE dataset across all evaluation metrics, establishing its superiority.

\subsection{Swin UNETR-empowered SuperRetina}
After observing promising results in our experiments by using large kernels to increase the receptive field of SuperRetina's encoder, we considered the possibility of further enhancing its performance by introducing a transformer-based encoder. The rationale behind this choice stems from the transformer's inherent capability to capture long-range dependencies, which could be advantageous for our task. However, training a transformer on a limited dataset presents significant challenges, which we will discuss in the subsequent paragraphs. 

To provide a comprehensive understanding of the modifications we made to SuperRetina's architecture, we first introduce the concepts of Swin Transformer~\cite{liu2021swin} and Swin UNETR~\cite{hatamizadeh2021swin}. These serve as foundational references for our modifications. Subsequently, we delve into explaining our specific architectural adjustments to SuperRetina and outline the unique approach we employed to train such a computationally intensive model on our small dataset.

\subsubsection{Swin transformer}
The primary factor contributing to the success of the Swin Transformer is its hierarchical design~\cite{liu2021swin}. Instead of treating the input image as a whole, it divides the image into non-overlapping patches, considering each patch as a token. The developers of Swin Transformer introduced the concept of shifted windows, where tokens only attend to a limited neighborhood of tokens, rather than attending to all tokens. By employing a multi-stage hierarchical architecture, the Swin Transformer effectively captures long-range dependencies while maintaining manageable computational complexity.

\subsubsection{Swin UNETR}
Swin UNETR was developed specifically for semantic segmentation tasks, merging the Swin Transformer and CNNs within a UNet-style architecture to address pixel-level segmentation~\cite{hatamizadeh2021swin}. The key benefit of the UNet-shaped architecture lies in its utilization of skip connections. In our study, we substituted the encoder of SuperRetina with the encoder from Swin UNETR.

\subsubsection{Reverse knowledge distillation}
Conventional approaches to knowledge distillation typically involve transferring knowledge from a larger model to a smaller one, resulting in improved accuracy for the lightweight model. However, in our case, where we aimed to utilize a model with long-term dependencies, training a transformer model with limited data, and even with self-supervision and transfer learning, yielded inferior performance compared to a CNN model. To enhance the performance of the larger model, we employed reverse knowledge distillation, a technique where we transferred knowledge from the small model (teacher) to the large model (student). We referred to this approach as reverse knowledge distillation. 

The loss function utilized in the reverse knowledge distillation strategy consisted of two components: first, the loss between the student network's predictions and the actual output, and second, the distillation loss between the student network and the teacher network outputs. During the training process, we followed the traditional steps while introducing an additional step in each iteration. In conjunction with the regular steps, we generated a keypoint heatmap using the teacher model and calculated the dice loss between the keypoint heatmaps of the student and teacher models. This loss was referred to as $l_{clf}^{RKD}$. Additionally, we implemented contrastive matching between the descriptors of the teacher and student models, known as the $l_{des}^{RKD}$. Both the detect RKD loss $l_{clf}^{RKD}$, and descriptor RKD loss $l_{des}^{RKD}$ were incorporated into the original detector and descriptor losses, respectively. Refer to the paper~\cite{liu2022semi} for more information on the original losses.
Equations~\ref{det_loss}, \ref{detector_cls_loss}, \ref{detector_loss_part1}, and \ref{detector_loss_part2} represent the detector loss of our Swin UNETR-empowered SuperRetina model with SuperRetina/LK-SuperRetina as a teacher.
The total loss of the detector~\ref{det_loss} is
\begin{equation}\label{det_loss}
    l_{det} = l_{clf}^{'} + l_{geo}
\end{equation}
\begin{equation}\label{detector_cls_loss}
    l_{clf}^{'} = l_{clf} + l_{clf}^{RKD} 
\end{equation}
\begin{equation}\label{detector_loss_part1}
    l_{clf}(I;Y) = 1-\frac{2. \sum_{i,j}(P\circ \tilde{Y})_{i,j}}{\sum_{i,j}(P \circ P)_{i,j}+\sum_{i,j}(\tilde{Y}\circ \tilde{Y})_{i,j}} 
\end{equation}
where $\tilde{Y}$ is the smoothed version of the binary ground truth labels $Y$ of the keypoints after blurring them with a 2D Gaussian.
\begin{equation}\label{detector_loss_part2}
    l_{clf}^{RKD}(I_S;I_T) = 1-\frac{2. \sum_{i,j}(P_S\circ P_T)_{i,j}}{\sum_{i,j}(P_S \circ P_S)_{i,j}+\sum_{i,j}(P_T\circ P_T)_{i,j}} 
\end{equation}
where $P_S$ stands for the keypoint heatmap of the student, and $P_T$ refers to the keypoint heatmap of the teacher model.\\ 
And $ l_{geo} $ is the dice loss between the output heatmap of the student model when the input is the image $ I $, and the inverse projection of the heatmap produced by the student when the input to it is the augmented version of the image $ I $ , $ I^{'} $.
Similarly, the new descriptor loss is a combination of the original descriptor loss and the reverse knowledge distillation loss as in~\ref{descriptor_loss}
\begin{equation}\label{descriptor_loss}
    l_{Des} = l_{des} + l_{des}^{RKD} 
\end{equation}
When feeding the image $ I $ and its augmented version $ I^{'} $ to the student network, we optain two tensors for the descriptors $ D $, and $ D^{'} $. For each keypoint $ (i,j) $ in the non-maximum supressed keypoint set $ \tilde{P} $, two distances are computed $ \Phi_{i,j}^{rand} $ between the descriptors of $ (i,j) $ in the set $ \tilde{P} $ and a random point from registered heatmap $H(\tilde{P})$. And $\Phi_{i,j}^{hard}$ the minimal distance. As \ref{desc_loss_part1} depicts
\begin{equation}\label{desc_loss_part1}
    l_{des}(I,H) = \sum_{(i,j)\in \tilde{P}}max(0,m+\Phi_{i,j}-\frac{1}{2}(\Phi_{i,j}^{rand}+\Phi_{i,j}^{hard}))   
\end{equation}
Similar to $l_{des}$, we compute $l_{des}^{RKD}$ between the descriptors generated when passing I to the student model, and the descriptors generated when passing I to the teacher model. For further details on the reverse knowledge distillation method and the loss functions, check our supplemental material~\cite{Authors14}.

\section{Experiments}
\label{sec:experiments}
We conducted a comprehensive evaluation of our proposed method by comparing its performance against various techniques in the retinal image matching task. Tab~\ref{tab:results1} presents a comprehensive comparison between our top-performing technique, Swin UNETR-empowered SuperRetina, and alternative approaches for retinal image matching, encompassing both traditional and deep learning-based methods. The results clearly demonstrate the utility of our proposed method, as it outperforms all other approaches.

The evaluation metrics consist of two aspects: the failure rate and the acceptance rate. The failure rate is determined by examining the number of matches between a query image, and its reference. A registration is considered failed if the number of matches is less than 4, which is the minimum required for estimating a homography, H. On the other hand, the acceptance rate is calculated for each query point in the query image. It involves computing the L2 distance between registered point and its corresponding reference point in the reference image. The median distance is defined as the median error (MEE) for each query image, with the maximum distance representing the maximum error (MAE). To be considered acceptable, MEE must be less than 20 and MAE must be less than 50. If these conditions are not met, the registration is deemed inaccurate. 

To assess the overall performance of a specific method, the area under receiver operating characteristic curve (AUC) is reported. AUC estimates the expectation of the acceptance rates with respect to the decision threshold. It reflects the performance across all methods. Additionally, AUC is computed separately for each category (Easy, Mod, Hard), and their mean (mAUC) is used as an overall measure.

In conclusion, the superior method is determined by having a higher acceptance rate or AUC, and lower rates of inaccuracies or failures.
To analyze the impact of different modifications to the encoder, varying kernel sizes of the large kernel-empowered SuperRetina, and diverse techniques for training the Swin Unetr encoder, we conducted ablation studies, the details listed below.

\begin{figure*}
  \centering
  \includegraphics[width=1.0\linewidth]{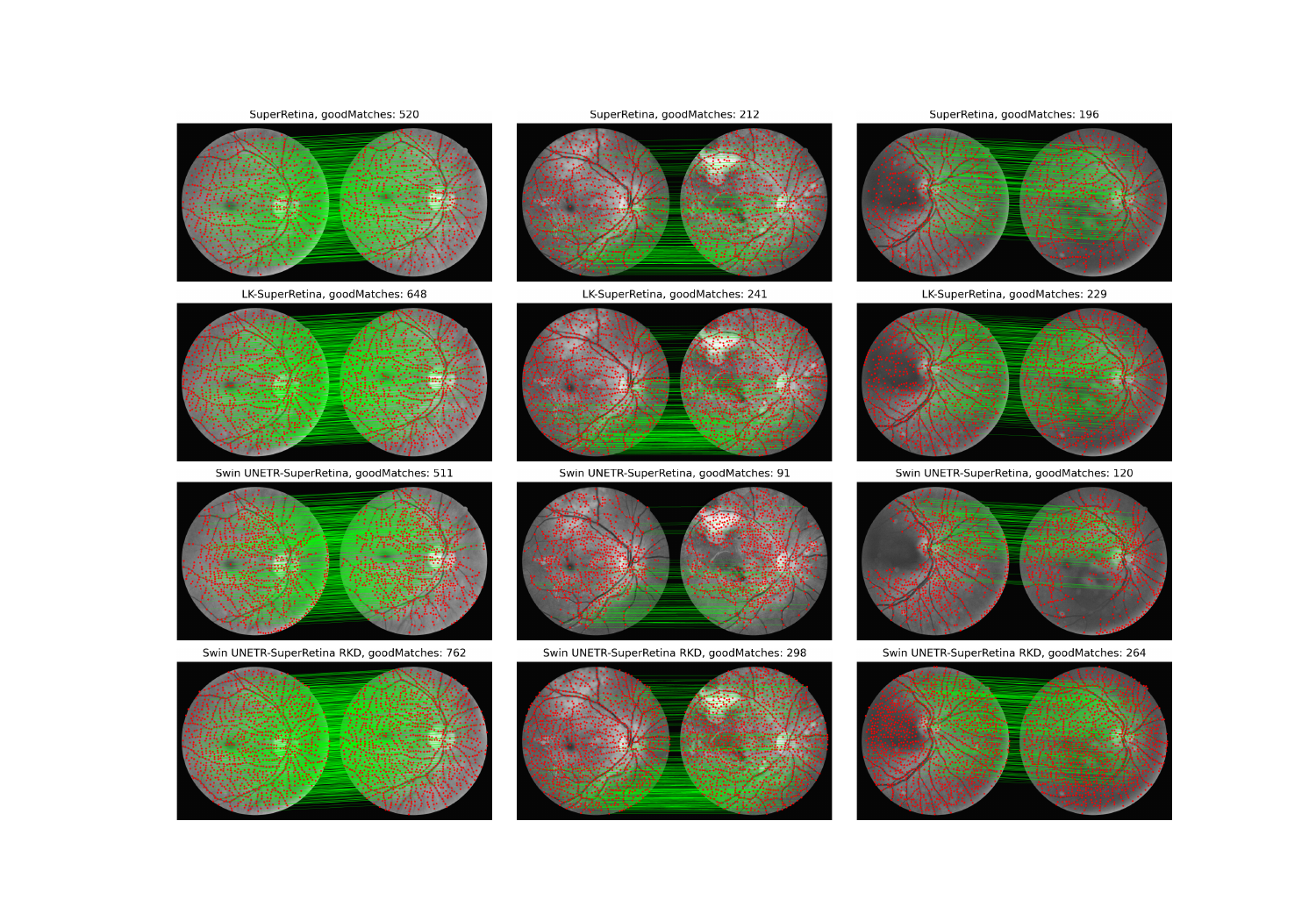}
  \vspace{-1cm}
  \caption{Performance comparison of our proposed methods on three example scenarios from FIRE dataset~\cite{hernandez2017fire}: class S (easy), class A (moderate), and class P (hard) from left to right. LK stands for large kernel, RKD refers to Reverse Knowledge Distillation with 50\% Dropout.}
  \label{matches}
\end{figure*}

\begin{table*}
  \centering
  \begin{tabular}{@{}lc@{}}
    \toprule
    Method & Failed\quad\quad Inaccurate\quad Acceptable\quad AUC-Easy AUC-Mod AUC-Hard mAUC \\
    \midrule
    SIFT, IJCV04~\cite{lowe2004distinctive} & \textbf{0.00\%}  \quad \quad 20.15\% \quad \quad 79.85\% \quad \quad 0.903 \quad \quad 0.474 \quad \quad 0.341 \quad \quad 0.573  \\
    PBO, ICIP10~\cite{oinonen2010identity}  & 0.75\% \quad \quad 28.36\% \quad \quad 70.89\% \quad \quad 0.844 \quad \quad 0.691 \quad \quad 0.122 \quad \quad 0.552 \\
    REMPE, JBHI20~\cite{hernandez2020rempe}  & \textbf{0.00\%} \quad \quad 02.99\% \quad \quad 97.01\% \quad \quad \textbf{0.958} \quad \quad 0.660 \quad \quad 0.542 \quad \quad 0.720\\
    SuperPoint, CVPRW18~\cite{detone2018superpoint} & \textbf{0.00\%} \quad \quad 05.22\% \quad \quad 94.78\% \quad \quad 0.882 \quad \quad 0.649 \quad \quad 0.490 \quad \quad 0.674\\
    GLAMpoints, ICCV19~\cite{truong2019glampoints} & \textbf{0.00\%} \quad \quad 07.46\% \quad \quad 92.54\% \quad \quad 0.850 \quad \quad 0.543 \quad \quad 0.474 \quad \quad 0.622 \\
    R2D2, NIPS19~\cite{revaud2019r2d2} & \textbf{0.00\%} \quad \quad 12.69\% \quad \quad 87.31\% \quad \quad 0.900 \quad \quad 0.517 \quad \quad 0.386 \quad \quad 0.601\\
    SuperGlue, CVPR20~\cite{sarlin2020superglue} & 0.75\% \quad \quad 03.73\% \quad \quad 95.52\% \quad \quad 0.885 \quad \quad 0.689 \quad \quad 0.488 \quad \quad 0.687 \\
    NCNet, TPAMI22~\cite{rocco2020ncnet} & \textbf{0.00\%} \quad \quad 37.31\% \quad \quad 62.69\% \quad \quad 0.588 \quad \quad 0.386 \quad \quad 0.077 \quad \quad 0.350 \\
    SuperRetina~\cite{liu2022semi:}& \textbf{0.00\%} \quad \quad 01.50\% \quad \quad 98.50\% \quad \quad 0.940 \quad \quad \textbf{0.783} \quad \quad 0.542 \quad \quad 0.755\\
    \textbf{Ours-1 (Large kernel-SuperRetina)} &\textbf{0.00\%}\quad\quad 00.75\% \quad \quad 99.25\% \quad\quad 0.942\quad\quad \textbf{0.783} \quad\quad \textbf{0.558} \quad\quad \textbf{0.761 }\\
    \textbf{Ours-2 (Swin UNETR-SuperRetina)} &\textbf{0.00\%}\quad \quad \textbf{00.00\%} \quad \quad \textbf{100.0\%} \quad \quad 0.935 \quad \quad 0.780 \quad \quad 0.550 \quad \quad 0.755\\
    \bottomrule
  \end{tabular}
  \caption{ A comparison among various techniques for retinal image matching, specifically focusing on the results obtained when testing the methods on the FIRE dataset~\cite{hernandez2017fire}. Our proposed method demonstrates superior performance when compared to both traditional and deep learning approaches. Ours-1 refers to large-kernel-empowered SuperRetina, while Ours-2 refers to Swin UNETR-empowered SuperRetina with SuperRetina as a teacher and drop out 50\%. In the table we provide the percentage values [\%] of failed, inaccurate, and acceptable.}
  \label{tab:results1}
\end{table*}

\subsection{Different kernel sizes}
Through conducting an ablation study focused on kernel size, we discovered that a combination of kernels with dimensions of $1 \times 1$, $3 \times 3$, and $5 \times 5$ yielded the most favorable outcomes of large kernel-empowered SuperRetina. See ablation studies in Table \ref{tab:results2}.

\subsection{Transfer learning}
In an attempt to address the difficulty of training a transformer model with limited data, we utilized transfer learning. To accomplish this, we gathered a substantial dataset of retinal images from online sources. This dataset was employed to train a Swin UNETR model on various tasks, including image inpainting and angle prediction. We then used pretrained encoder weights as initial weights for the SuperRetina's encoder. Tab.~\ref{tab:results2} shows the results of using a pretrained Swin UNETR as a backbone of SuperRetina. Unfortunately, this model outperforms other models for one specific evaluation metric, namely AUC-Mod.

\subsection{Reverse knowledge distillation}
Transformers, as highlighted in the study by Dosovitskiy et al. (2020) \cite{dosovitskiy2020image}, have a high demand for extensive training data and tend to perform less effectively than CNNs when dealing with limited data. Reverse knowledge distillation refers to using the knowledge obtained by a smaller model, e.g. a CNN, to train a larger model, e.g. a Transformers. Typically, the knowledge of a larger model is used to train a smaller model in knowledge distillation, as discussed in works such as Chen et al. (2022) \cite{chen2022dearkd}, Touvron et al. (2021) \cite{touvron2021training}, and Hinton et al. (2015) \cite{hinton2015distilling}. Within our research, we regard the CNN as the "teacher" model, previously trained for the keypoint detection task. The objective is to transfer the CNN's knowledge and generalization abilities to a transformer model, referred to as the "student" model. The distillation process entails training the student model to imitate the behavior of the teacher model. This typically involves employing the output probabilities or feature representations of the teacher model as soft targets during the student model's training. By mimicking the teacher's predictions, the student model effectively captures the teacher model's knowledge and decision-making process. We anticipated that by distilling knowledge from a CNN to a transformer, the transformer model could potentially benefit from both the CNN's local feature extraction capabilities and the transformer's ability to model long-range dependencies. However, our experimental results indicate that even after knowledge distillation, the transformer model performed inferiorly compared to our expectations. See Table \ref{tab:results2}. To resolve this problem, as shown in Table~\ref{tab:results2}, we incorporated a 50\% dropout, which significantly enhanced the performance of the Swin UNETR-empowered SuperRetina, enabling it to achieve 100\% accuracy on the testing dataset. This improvement arises from the network's enhanced generalization capability on testing data by mitigating overfitting on the training data combined with reverse knowledge distillation. We conclude that regularization strategies, such as dropout, will be critical to reverse knowledge distillation. By using the dropout trick, we observed an enhancement in the student model's generalization capability, enabling it to outperform its teacher model on the testing dataset.
See Fig.~\ref{matches} for a visual comparison between our proposed methods.

\begin{table}
  \centering
  \Rotatebox{90}{%
  \begin{tabular}{@{}lr@{}}
    \toprule
    Method & Failed\quad \quad Inaccurate\quad Acceptable\quad AUC-Easy AUC-Mod AUC-Hard mAUC \\
    \midrule
    SuperRetina~\cite{liu2022semi}, KS $3 \times 3$& \textbf{0.00\%} \quad \quad 01.50\% \quad \quad 98.50\% \quad \quad 0.940 \quad \quad \textbf{0.783} \quad \quad 0.542 \quad \quad 0.755\\
    LK-SuperRetina, KS $1 \times 1, 3 \times 3, 5 \times 5$ &\textbf{0.00\%} \quad\quad 00.75\% \quad \quad 99.25\% \quad\quad 0.942\quad\quad \textbf{0.783} \quad\quad 0.558 \quad\quad \textbf{0.761}\\
    LK-SuperRetina, KS $1 \times 1, 3 \times 3, 5 \times 5, 7 \times 7$ & \textbf{0.00\%} \quad \quad 02.25\% \quad \quad 97.74\%  \quad \quad 0.925 \quad \quad 0.717 \quad\quad 0.502 \quad \quad 0.714\\
    Swin UNETR-SuperRetina, Trained from scratch & \textbf{0.00\%} \quad \quad 16.55\% \quad \quad 83.45\% \quad \quad 0.891 \quad \quad 0.649 \quad \quad 0.318 \quad \quad 0.619\\
    Swin UNETR-SuperRetina, SuperRetina as teacher w/o dropout (DO) & \textbf{0.00\%} \quad \quad 01.5\% \quad \quad 98.50\% \quad \quad \textbf{0.947} \quad \quad 0.769 \quad \quad 0.549 \quad \quad 0.755\\
    Swin UNETR-SuperRetina, SuperRetina as teacher,  DO 50\% & \textbf{0.00\%} \quad \quad \textbf{00.00\%} \quad \quad \textbf{100.0\%}\quad \quad 0.935 \quad \quad 0.780 \quad \quad 0.550 \quad \quad 0.755\\
    Swin UNETR-SuperRetina, LK-SuperRetina as teacher,  DO 50\%& \textbf{0.00\%}\quad \quad00.75\%\quad \quad 99.25\%\quad \quad 0.914 \quad \quad 0.774 \quad \quad 0.558 \quad \quad 0.749 \\
    Pretrained Swin UNETR-SuperRet., LK-SuperRet. as teacher, DO 50\%&\textbf{0.00\%} \quad \quad 00.75\% \quad \quad 99.25\% \quad \quad 0.928 \quad \quad 0.774 \quad \quad \textbf{0.559} \quad \quad 0.754\\
    \bottomrule
  \end{tabular}
  }
  \caption{Ablation testing on FIRE dataset~\cite{hernandez2017fire} (KS is kernel size).}
  \label{tab:results2}
\end{table}

\section{Conclusion}
\label{sec:conclusion}

In conclusion, our research aimed to advance retinal image matching through a novel approach utilizing a semi-supervised learning framework. We successfully enhanced the SuperRetina model and investigated the relevance of CNN-based methods compared to vision transformers. By introducing tailored architectural modifications, we surpassed the performance of state-of-the-art keypoint detection architectures in retinal images. Surprisingly, we found that strategic exploitation of the CNN encoder's receptive field was sufficient to capture accurate and discriminative keypoints, eliminating the need for long-range dependencies. This modification to the SuperRetina architecture achieved state-of-the-art performance.

We also explored training methodologies for large models with limited data and employed reverse knowledge distillation by using a smaller CNN model to guide a larger transformer during its training. This approach proved effective in overcoming the challenge of limited data, yielding 100\% accuracy on the testing dataset. Moreover, we contributed to the research community by providing a public dataset with annotations for retinal image detection and matching, enabling the development of algorithms for various retinal image applications.

{\small
\bibliographystyle{ieee_fullname}
\bibliography{PaperForReview}

\begin{thebibliography}{10}\itemsep=-1pt

\bibitem{e-ophtha}
e-ophtha dataset.
\newblock \url{https://www.adcis.net/en/third-party/e-ophtha/}.
\newblock Accessed: 2023-04-20.

\bibitem{KaggleDiabeticRetinopathy}
Kaggle diabetic retinopathy detection competition dataset.
\newblock
  \url{https://www.kaggle.com/competitions/diabetic-retinopathy-detection/data?select=train.zip.005}.
\newblock Accessed: 01/06/20.

\bibitem{KaggleGlaucomaDatasets}
Kaggle glaucoma datasets.
\newblock \url{https://www.kaggle.com/datasets/arnavjain1/glaucoma-datasets}.
\newblock Accessed: 01/06/2023.

\bibitem{RetinalDiseaseClassification}
Kaggle retinal disease classification dataset.
\newblock
  \url{https://www.kaggle.com/datasets/andrewmvd/retinal-disease-classification}.
\newblock Accessed: 01/06/2023.

\bibitem{RetinalDisease}
Retinal disease classification dataset.
\newblock
  \url{https://www.kaggle.com/datasets/andrewmvd/retinal-disease-classification}.
\newblock Accessed: 2023-04-20.

\bibitem{addison2015low}
Jimmy Addison~Lee, Jun Cheng, Beng Hai~Lee, Ee Ping~Ong, Guozhen Xu, Damon Wing
  Kee~Wong, Jiang Liu, Augustinus Laude, and Tock Han~Lim.
\newblock A low-dimensional step pattern analysis algorithm with application to
  multimodal retinal image registration.
\newblock In {\em Proceedings of the IEEE Conference on Computer Vision and
  Pattern Recognition}, pages 1046--1053, 2015.

\bibitem{Authors14}
Anonymous.
\newblock Supplementary material of the paper reverse knowledge distillation:
  Training a large model using a small one for retinal image matching on
  limited data, 2023.
\newblock Supplied as supplemental material {\tt supplemental802.pdf}.

\bibitem{balntas2017hpatches}
Vassileios Balntas, Karel Lenc, Andrea Vedaldi, and Krystian Mikolajczyk.
\newblock Hpatches: A benchmark and evaluation of handcrafted and learned local
  descriptors.
\newblock In {\em Proceedings of the IEEE conference on computer vision and
  pattern recognition}, pages 5173--5182, 2017.

\bibitem{bay2006surf}
Herbert Bay, Tinne Tuytelaars, and Luc Van~Gool.
\newblock Surf: Speeded up robust features.
\newblock {\em Lecture notes in computer science}, 3951:404--417, 2006.

\bibitem{chen2022dearkd}
Xianing Chen, Qiong Cao, Yujie Zhong, Jing Zhang, Shenghua Gao, and Dacheng
  Tao.
\newblock Dearkd: data-efficient early knowledge distillation for vision
  transformers.
\newblock In {\em Proceedings of the IEEE/CVF Conference on Computer Vision and
  Pattern Recognition}, pages 12052--12062, 2022.

\bibitem{christiansen2019unsuperpoint}
Peter~Hviid Christiansen, Mikkel~Fly Kragh, Yury Brodskiy, and Henrik Karstoft.
\newblock Unsuperpoint: End-to-end unsupervised interest point detector and
  descriptor.
\newblock {\em arXiv preprint arXiv:1907.04011}, 2019.

\bibitem{deng2009imagenet}
Jia Deng, Wei Dong, Richard Socher, Li-Jia Li, Kai Li, and Li Fei-Fei.
\newblock Imagenet: A large-scale hierarchical image database.
\newblock In {\em 2009 IEEE conference on computer vision and pattern
  recognition}, pages 248--255. Ieee, 2009.

\bibitem{detone2018superpoint}
Daniel DeTone, Tomasz Malisiewicz, and Andrew Rabinovich.
\newblock Superpoint: Self-supervised interest point detection and description.
\newblock In {\em Proceedings of the IEEE conference on computer vision and
  pattern recognition workshops}, pages 224--236, 2018.

\bibitem{dosovitskiy2020image}
Alexey Dosovitskiy, Lucas Beyer, Alexander Kolesnikov, Dirk Weissenborn,
  Xiaohua Zhai, Thomas Unterthiner, Mostafa Dehghani, Matthias Minderer, Georg
  Heigold, Sylvain Gelly, et~al.
\newblock An image is worth 16x16 words: Transformers for image recognition at
  scale.
\newblock {\em arXiv preprint arXiv:2010.11929}, 2020.

\bibitem{medal2023}
Nihar Gupte, Sahar Almahfouz~Nasser, Prateek Garg, Keshav Singhal, Tanmay Jain,
  Aditya, Ravi Kumar, and Amit Sethi.
\newblock {MeDAL-Retina}.
\newblock
  \url{https://www.dropbox.com/sh/o8q84e2eg54ay3d/AADiAkNr6bFQDoFaKeEjpYtra?dl=0},
  2023.
\newblock Dataset.

\bibitem{habib2023knowledge}
Gousia Habib, Tausifa~Jan Saleem, and Brejesh Lall.
\newblock Knowledge distillation in vision transformers: A critical review.
\newblock {\em arXiv preprint arXiv:2302.02108}, 2023.

\bibitem{hatamizadeh2021swin}
Ali Hatamizadeh, Vishwesh Nath, Yucheng Tang, Dong Yang, Holger~R Roth, and
  Daguang Xu.
\newblock Swin unetr: Swin transformers for semantic segmentation of brain
  tumors in mri images.
\newblock In {\em International MICCAI Brainlesion Workshop}, pages 272--284.
  Springer, 2021.

\bibitem{hernandez2020rempe}
Carlos Hernandez-Matas, Xenophon Zabulis, and Antonis~A Argyros.
\newblock Rempe: Registration of retinal images through eye modelling and pose
  estimation.
\newblock {\em IEEE journal of biomedical and health informatics},
  24(12):3362--3373, 2020.

\bibitem{hernandez2017fire}
Carlos Hernandez-Matas, Xenophon Zabulis, Areti Triantafyllou, Panagiota
  Anyfanti, Stella Douma, and Antonis~A Argyros.
\newblock Fire: fundus image registration dataset.
\newblock {\em Modeling and Artificial Intelligence in Ophthalmology},
  1(4):16--28, 2017.

\bibitem{hinton2015distilling}
Geoffrey Hinton, Oriol Vinyals, and Jeff Dean.
\newblock Distilling the knowledge in a neural network.
\newblock {\em arXiv preprint arXiv:1503.02531}, 2015.

\bibitem{jia2022u}
Xi Jia, Joseph Bartlett, Tianyang Zhang, Wenqi Lu, Zhaowen Qiu, and Jinming
  Duan.
\newblock U-net vs transformer: Is u-net outdated in medical image
  registration?
\newblock In {\em Machine Learning in Medical Imaging: 13th International
  Workshop, MLMI 2022, Held in Conjunction with MICCAI 2022, Singapore,
  September 18, 2022, Proceedings}, pages 151--160. Springer, 2022.

\bibitem{liu2022semi}
Jiazhen Liu, Xirong Li, Qijie Wei, Jie Xu, and Dayong Ding.
\newblock Semi-supervised keypoint detector and descriptor for retinal image
  matching.
\newblock In {\em Computer Vision--ECCV 2022: 17th European Conference, Tel
  Aviv, Israel, October 23--27, 2022, Proceedings, Part XXI}, pages 593--609.
  Springer, 2022.

\bibitem{liu2022semi:}
Jiazhen Liu, Xirong Li, Qijie Wei, Jie Xu, and Dayong Ding.
\newblock Semi-supervised keypoint detector and descriptor for retinal image
  matching.
\newblock In {\em Computer Vision--ECCV 2022: 17th European Conference, Tel
  Aviv, Israel, October 23--27, 2022, Proceedings, Part XXI}, pages 593--609.
  Springer, 2022.

\bibitem{liu2021swin}
Ze Liu, Yutong Lin, Yue Cao, Han Hu, Yixuan Wei, Zheng Zhang, Stephen Lin, and
  Baining Guo.
\newblock Swin transformer: Hierarchical vision transformer using shifted
  windows.
\newblock In {\em Proceedings of the IEEE/CVF international conference on
  computer vision}, pages 10012--10022, 2021.

\bibitem{lowe2004distinctive}
David~G Lowe.
\newblock Distinctive image features from scale-invariant keypoints.
\newblock {\em International journal of computer vision}, 60:91--110, 2004.

\bibitem{oinonen2010identity}
Hannu Oinonen, Heikki Forsvik, Pekka Ruusuvuori, Olli Yli-Harja, Ville Voipio,
  and Heikki Huttunen.
\newblock Identity verification based on vessel matching from fundus images.
\newblock In {\em 2010 IEEE International Conference on Image Processing},
  pages 4089--4092. IEEE, 2010.

\bibitem{fire_dataset}
{Papers With Code}.
\newblock Fire: Framework for information retrieval evaluation.
\newblock \url{https://paperswithcode.com/dataset/fire}, Accessed: 2023.

\bibitem{revaud2019r2d2}
Jerome Revaud, Philippe Weinzaepfel, C{\'e}sar De~Souza, Noe Pion, Gabriela
  Csurka, Yohann Cabon, and Martin Humenberger.
\newblock R2d2: repeatable and reliable detector and descriptor.
\newblock {\em arXiv preprint arXiv:1906.06195}, 2019.

\bibitem{rocco2020ncnet}
Ignacio Rocco, Mircea Cimpoi, Relja Arandjelovi{\'c}, Akihiko Torii, Tomas
  Pajdla, and Josef Sivic.
\newblock Ncnet: Neighbourhood consensus networks for estimating image
  correspondences.
\newblock {\em IEEE Transactions on Pattern Analysis and Machine Intelligence},
  44(2):1020--1034, 2020.

\bibitem{ronneberger2015u}
Olaf Ronneberger, Philipp Fischer, and Thomas Brox.
\newblock U-net: Convolutional networks for biomedical image segmentation.
\newblock In {\em Medical Image Computing and Computer-Assisted
  Intervention--MICCAI 2015: 18th International Conference, Munich, Germany,
  October 5-9, 2015, Proceedings, Part III 18}, pages 234--241. Springer, 2015.

\bibitem{rublee2011orb}
Ethan Rublee, Vincent Rabaud, Kurt Konolige, and Gary Bradski.
\newblock Orb: An efficient alternative to sift or surf.
\newblock In {\em 2011 International conference on computer vision}, pages
  2564--2571. Ieee, 2011.

\bibitem{ryu2011formula}
JB Ryu, CG Lee, and HH Park.
\newblock Formula for harris corner detector.
\newblock {\em Electronics letters}, 47(3):1, 2011.

\bibitem{sarhan2021transfer}
Abdullah Sarhan, Jon Rokne, Reda Alhajj, and Andrew Crichton.
\newblock Transfer learning through weighted loss function and group
  normalization for vessel segmentation from retinal images.
\newblock In {\em 2020 25th International Conference on Pattern Recognition
  (ICPR)}, pages 9211--9218. IEEE, 2021.

\bibitem{sarlin2020superglue}
Paul-Edouard Sarlin, Daniel DeTone, Tomasz Malisiewicz, and Andrew Rabinovich.
\newblock Superglue: Learning feature matching with graph neural networks.
\newblock In {\em Proceedings of the IEEE/CVF conference on computer vision and
  pattern recognition}, pages 4938--4947, 2020.

\bibitem{steiner2021train}
Andreas Steiner, Alexander Kolesnikov, Xiaohua Zhai, Ross Wightman, Jakob
  Uszkoreit, and Lucas Beyer.
\newblock How to train your vit? data, augmentation, and regularization in
  vision transformers.
\newblock {\em arXiv preprint arXiv:2106.10270}, 2021.

\bibitem{touvron2021training}
Hugo Touvron, Matthieu Cord, Matthijs Douze, Francisco Massa, Alexandre
  Sablayrolles, and Herv{\'e} J{\'e}gou.
\newblock Training data-efficient image transformers \& distillation through
  attention.
\newblock In {\em International conference on machine learning}, pages
  10347--10357. PMLR, 2021.

\bibitem{truong2019glampoints}
Prune Truong, Stefanos Apostolopoulos, Agata Mosinska, Samuel Stucky, Carlos
  Ciller, and Sandro~De Zanet.
\newblock Glampoints: Greedily learned accurate match points.
\newblock In {\em Proceedings of the IEEE/CVF International Conference on
  Computer Vision}, pages 10732--10741, 2019.

\bibitem{weiss2016survey}
Karl Weiss, Taghi~M Khoshgoftaar, and DingDing Wang.
\newblock A survey of transfer learning.
\newblock {\em Journal of Big data}, 3(1):1--40, 2016.

\bibitem{yu2022metaformer}
Weihao Yu, Mi Luo, Pan Zhou, Chenyang Si, Yichen Zhou, Xinchao Wang, Jiashi
  Feng, and Shuicheng Yan.
\newblock Metaformer is actually what you need for vision.
\newblock In {\em Proceedings of the IEEE/CVF conference on computer vision and
  pattern recognition}, pages 10819--10829, 2022.

\end{thebibliography}
}

\end{document}